\documentclass[11pt]{article}
\usepackage{eamt24}
\usepackage{times}
\usepackage{url}
\usepackage{latexsym}
\usepackage[small,bf]{caption} %
\usepackage{booktabs}
\usepackage{todonotes}
\usepackage{amsmath}
\usepackage{xcolor}
\usepackage{pifont}

\newcommand{\aal}[0]{\textsuperscript{\ding{100}}}
\newcommand{\aals}[0]{\ding{100}}
\newcommand{\rug}[0]{\textsuperscript{\ding{80}}}

\title{Towards Tailored Recovery of Lexical Diversity in Literary \\ Machine Translation}

 \author{Esther Ploeger\aals \hspace{0.8em}
  Huiyuan Lai\rug \hspace{0.8em}
  Rik van Noord\rug \hspace{0.8em}
  Antonio Toral\rug  \\
  \aal Department of Computer Science, Aalborg University, Denmark \\ 
  \rug  CLCG, University of Groningen, The Netherlands \\ 
  \texttt{espl@cs.aau.dk} \quad
  \texttt{\{h.lai,r.i.k.van.noord,a.toral.ruiz\}@rug.nl}
  }

\date{}

\begin{document}
\maketitle

\begin{abstract}
Machine translations are found to be lexically poorer than human translations.
The loss of lexical diversity through MT poses an issue in the automatic translation of literature, where it matters not only \textit{what} is written, but also \textit{how} it is written.
Current methods for increasing lexical diversity in MT are rigid.
Yet, as we demonstrate, the %
degree of lexical diversity can vary considerably
across different novels.
Thus, rather than aiming for the rigid %
\textit{increase}
of lexical diversity, we reframe the task as \textit{recovering} what is lost in the machine translation process.
We propose a novel approach that consists of reranking translation candidates with a classifier that distinguishes between original and translated text.
We evaluate our approach on 31 English-to-Dutch book translations, and find that, for certain books, our approach retrieves lexical diversity scores that are close to human translation. %
\end{abstract}

\section{Introduction}

With the introduction of neural machine translation (NMT), the performance of high-resource automatic translation has improved substantially.
Especially since the introduction of the Transformer architecture \cite{vaswani2017attention}, state-of-the-art NMT systems have outperformed previous approaches considerably \cite{lakew-etal-2018-comparison}, with some works even claiming human parity \cite{popel2020transforming}. %
\begin{figure}[t]
    \centering
    \includegraphics[width=0.48\textwidth]{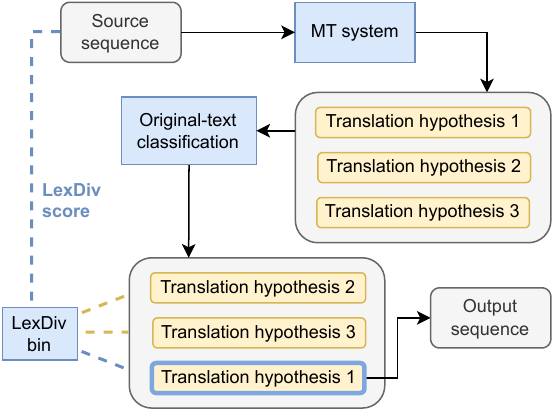}
    \caption{Reranking translation hypotheses based on the probability they are originally written in the target language, where the chosen rank is based on the lexical diversity score of the original book, and could be lower than the most lexically diverse option.}
    \label{fig:overview}
\end{figure}
However, these claims are based mostly on accuracy and fluency measures, while style is often overlooked. In fact, according to expert evaluation, machine translation (MT) did actually not reach human parity \cite{toral-etal-2018-attaining,fischer-laubli-2020-whats}.
For instance, MT models have been found to exacerbate linguistic patterns that occur frequently, while underrepresenting patterns that are found less commonly \cite{vanmassenhove2019lost}. As a result, automatically translated texts are found to be lexically poorer than human translations (HT). This `artificially impoverished language' has previously been referred to as \textit{machine translationese} \cite{vanmassenhove-etal-2021-machine}.

In this paper, we focus on the translation of novels. Contrary to technical domains, where meaning preservation is the main criterion for acceptable translations, literary translations have the additional criterion of style.
This is because apart from meaning preservation (\textit{what} is written), maintaining a certain reading experience (\textit{how} it is written) is vital for novels \cite{toral2015machine}.
Importantly however,  writing style (and linguistic complexity) can vary considerably between books. Some books contain repetitive language use, while others are written in embellished language (see Section \ref{sec:why}).
Current approaches that aim to mitigate the loss of lexical diversity do not accommodate this. State-of-the-art previous work \cite{freitag-etal-2019-ape,freitag-etal-2022-natural} increases lexical diversity in a rigid way, not allowing for flexibility at inference time.

\paragraph{Contributions}
\textit{(i)} We show that lexical diversity varies 
considerably across
books, and argue that this should be taken into account in MT;
\textit{(ii)} We introduce a novel flexible method for recovering
lexical diversity in MT, informed by the diversity of the original.
\textit{(ii)} We evaluate our method on 31 English novels which are translated to Dutch, and find that our approach is effective when it comes to book-tailored promotion of lexical diversity.

\section{Related Work}

\paragraph{Literary MT}
NMT has been argued to hold potential for literary texts, for instance in assisting professional translators or improving the immediate accessibility of untranslated foreign language books \cite{matusov-2019-challenges}.
However, MT has been shown to decrease lexical diversity \cite{vanmassenhove2019lost,vanmassenhove-etal-2021-machine}.
This is an issue, because literary works can be viewed as a special domain in translation. Typically, literary translators are expected to preserve not only literal elements from the source, such as the plot, but also some sense of creative value \cite{jorge-braga-riera-2022}. In other words, a goal of literary translation could be to recreate the `aesthetic intentions or effects' that are possibly present in the source book \cite{delabastita2011literary}.
Such `aesthetic intentions' can for instance be voice and metaphor, but also repetition \cite{wright2016literary}.
Repetitive use of language is commonly a conscious choice by the writer, and has a function, such as drawing attention or establishing a pattern \cite{boase2011critical}.
Given that lexical diversity can be an intentional writing choice, it should be apparent that an approach that aims at recovering lexical diversity in MT should not be boundless.
Therefore, it is our aim to inform recovery with the degree of relative lexical diversity of the source text.

\paragraph{Machine \textit{Translationese}}
Following  recommendations from Jiménez-Crespo \shortcite{jimenez-crespo-2023-translationese}, we will largely refrain from using the term \textit{translationese} in the rest of this paper.
However, it is important to note that previous work that aims to increase lexical diversity in MT has mostly been framed as part of `machine translationese' reduction \cite{freitag-etal-2019-ape,freitag-etal-2022-natural,dutta-chowdhury-etal-2022-towards,jalota-etal-2023-translating}. %
Translations have been found to differ from original texts in a number of ways.
For one, \newcite{baker1993corpus} argues that human translations into a language tend to be lexically simpler than text originally written in that language.
Automatic classification approaches have been effective in detecting this difference \cite{baroni2006,koppel-ordan-2011-translationese,volansky2015features,rabinovich-wintner-2015-unsupervised,pylypenko-etal-2021-comparing}. 
More recently, work has investigated linguistic differences between MT and HT \cite{van-der-werff-etal-2022-automatic}.
Thus, it seems that modelling characteristics of original versus translated texts has a direct link to lexical diversity.
Previous work \cite{freitag-etal-2022-natural}  leveraged these detectable differences in their approach to increase the naturalness of output translations.
We take inspiration from their lexical diversity evaluation methods, and implement their method as a baseline.

\paragraph{Reranking Methods}
Reranking hypotheses in text generation originated before the age of neural paradigms \cite{shen2004discriminative,collins2005discriminative}.
In essence, reranking entails re-ordering the set of candidate outputs according to some criterion, with the aim of providing a final output that adheres better to that criterion.
Such methods have been applied for various tasks, such as  summarization \cite{liu-liu-2021-simcls} and semantic parsing \cite{yin-neubig-2019-reranking}. In machine translation, previous approaches include discriminative reranking \cite{lee-etal-2021-discriminative} and reranking with energy-based models \cite{arcadinho-etal-2022-t5ql}.

\section{Why Recover Rather Than Increase Lexical Diversity?} %
\label{sec:why}

\begin{figure*}
    \centering
    \includegraphics[width=0.4\textwidth]{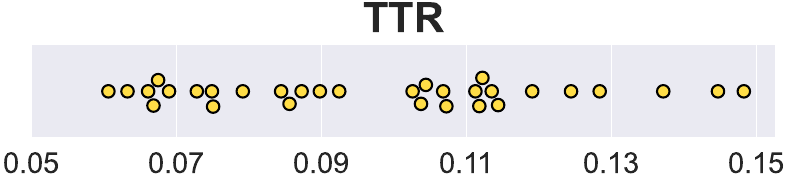}
    \hspace{3em}
    \vspace{0.25em}
    \includegraphics[width=0.4\textwidth]{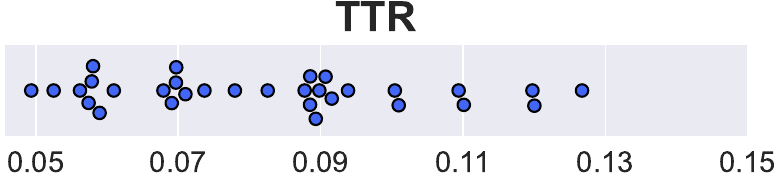}
    \vspace{0.25em}
    \includegraphics[width=0.4\textwidth]{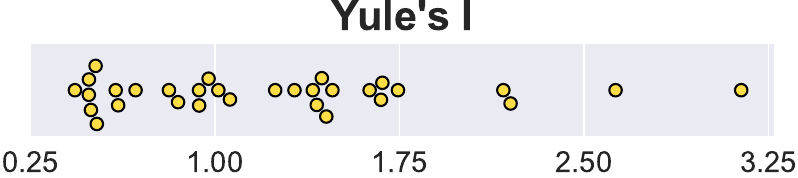}
    \hspace{3em}
    \includegraphics[width=0.4\textwidth]{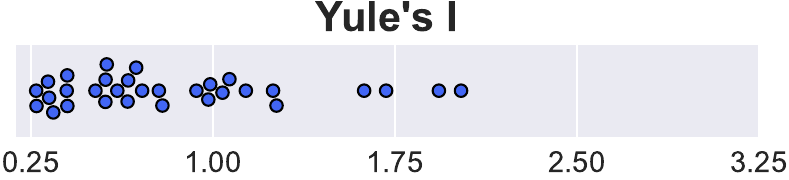}
        \includegraphics[width=0.4\textwidth]{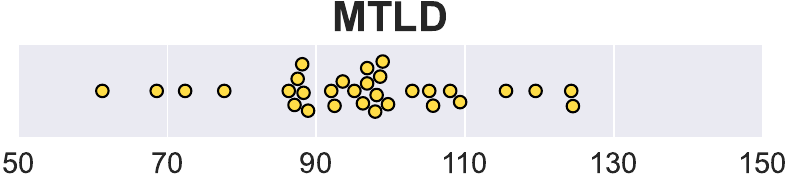}
    \hspace{3em}
    \includegraphics[width=0.4\textwidth]{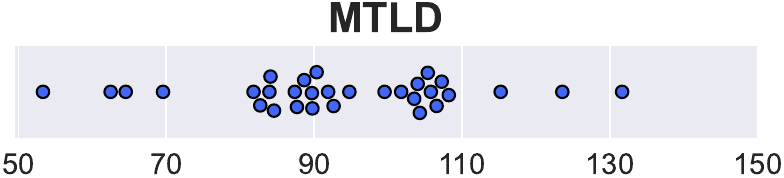}
    \caption{Range and spread of lexical diversity metrics for HT (left, yellow) and original English (right, blue).}
    \label{fig:range_and_spread}
\end{figure*}

In this paper, we argue for tailored recovery of lexical diversity.
In this section, we first discuss support for this idea from the field of literary studies. Then, we provide empirical evidence by applying lexical diversity metrics to our test set.

\subsection{Theoretical Support}
Previous work on writing style in novels acknowledges that some books exhibit more lexical diversity than others.
As an example, \newcite{heaton1970style}
finds that  no word in the original (i.e. English) version of in \textit{The Old man and the Sea} by Ernest Hemingway contains more than six syllables. Additionally, Hemingway tends to stick to particular words, even when there are more diverse options: in 184 situations of direct speech, he chooses to use the word `said' 170 times instead of for example `asked', `remarked', `noticed' or `yelled'.
An example from the other end of the spectrum is James Joyce's \textit{Ulysses}.
This work is known for its experimental techniques and unorthodox language use.  \newcite{trotta2014creativity} illustrates this by highlighting Joyce's use of neologisms, such as `He \textit{smellsipped} the cordial juice' and `Davy Byrne \textit{smiledyawnednodded} all in one'. Moreover, Joyce repeatedly uses non-verbs as verbs, like in `I am \textit{almosting} it.' and even writes long sequences in unconventional spelling (\textit{Ahbeesee defeegee kelomen opeecue rustyouvee doubleyou}. Boys are they?').
These examples make it clear that books can be written with vastly different `aesthetic intentions'. Thus, for preserving these intentions, MT approaches should not render them equally diverse in terms of lexicon.

\subsection{Empirical Support}

We empirically verify whether these findings hold for our data specifically, by estimating the lexical diversity of the 31 books in our test set, which we introduce in Section \ref{sec:vanilla-mt}.
We calculate three measures of lexical variety (type-token ratio; TTR, Yule's I \cite{yule1944statistical}, and MTLD \cite{mccarthy2005assessment}) for each book in our test set.
We further elaborate on these metrics in Section \ref{sec:eval}.
Next, we apply the same metrics to the human translations of those same books.
Figure \ref{fig:range_and_spread} shows that there is indeed a wide range of diversity across books, for both HT and original text.
For example, in both settings, we find that the highest MTLD value is almost two times as large as the lowest.
This emphasises why it is not our aim to generate the highest possible lexical diversity for every book.
While we observe similar ranges and distributions in HT vs. original, the HT metrics are slightly higher.
However, this does not necessarily mean that HT contains more embellished language.
We note that the languages in our study, Dutch and English, are relatively similar (both in terms of genealogy and linguistic typology), but they differ in ways that can influence diversity metrics. For instance, Dutch contains compound nouns while English does not, making a higher TTR for Dutch more likely.

This discrepancy means that we cannot compare our Dutch MT to the original English book diversity directly. Instead, here we compare MT with HT.
To verify whether this is sensible, we assess the relationship between HT and the English originals, by computing Pearson's correlation on the corresponding diversity metrics. The results are listed in Table~\ref{tab:lexdiv-corr}, 
and the corresponding
regression plots are found in Appendix B. We observe strong correlations that are all statistically significant. This is important, because as the source diversity is a reliable indicator of HT diversity, it makes sense to use the source scores to approach HT (see Section \ref{sec:reranking}).

\begin{table}[h]
    \centering
    \begin{tabular}{lcc}
    \toprule
         \textbf{Metric} & \textbf{Correlation coefficient} & \textbf{\textit{p}-value}\\
     \midrule
        TTR & 0.971 & $<$ 0.00001 \\
        Yule's I & 0.929 & $<$ 0.00001 \\
        MTLD  &  0.953 & $<$ 0.00001 \\
     \bottomrule
    \end{tabular}
    \caption{Pearson correlation coefficients for HT and OR lexdiv metrics, rounded to three decimals.}
    \label{tab:lexdiv-corr}
\end{table}

\section{Reranking Method}
\label{sec:reranking}

As illustrated in Figure \ref{fig:overview}, our approach consists of two parts: hypothesis generation and hypothesis reranking.
Firstly, we generate the \textit{n} best translation candidates for each source sentence in the test set with a vanilla domain-specific MT system (Section \ref{sec:vanilla-mt}). Note that we decode all books separately, instead of concatenating all test set books.
Then, for each book, we apply a classifier (Section \ref{sec:cls}) to the translation hypotheses and, through a softmax layer, obtain the probability for each candidate that it is an original Dutch sequence. 
Based on these probabilities, we rerank the translation candidates.
In order to obtain the (expected) most lexically rich candidate, we would then choose the rank with the highest original-text probability.
However, note that this simple approach is flexible in the sense that, instead of choosing the most original-like option, we have the option to choose a lower original-text rank.

We leverage this flexibility for tailoring rank selection to the lexical diversity of the original English book.
First, for each original book, we calculate a \textit{LexDiv} score, which consists of the average of the normalized TTR, Yule's I and MTLD scores (see Section \ref{sec:eval}).
Then, we bin the books according to their \textit{LexDiv} score, relative to the total distribution.
That is, given a list that is sorted based on \textit{LexDiv}, we categorize these into groups, where the number of groups depends on the number of \textit{nbest} candidates in decoding.
For example, for $n=5$, we bin the books into 5 different groups of 6 books (adding any remainders into the last bin). The bin per book corresponds to the original-text rank that is selected.
As such, the selected rank for each book depends on the lexical diversity of its source, relative to the other books.
Reranking translation candidates is a suitable solution to our task, because it accomodates flexibility, which is tunable at inference time. There is no need to train a separate model per diversity setting, saving computational expenses. 
Additionally, our approach is model-agnostic: reranking can be applied to any MT model that can generate multiple translation candidates.

\section{Experimental set-up}

\subsection{Vanilla MT System}
\label{sec:vanilla-mt}

\paragraph{Data}
We use the dataset by %
\newcite{toral-cranenburgh-nutters-2024}, which contains 531 books that were originally written in English and manually translated into Dutch.
We use 495 books for training, 5 for development and 31 as a test set.
The genres of the books vary: they include literary fiction, popular fiction, non-fiction and children's books from over 100 authors.
We do not make a distinction between literary and `unliterary' novels, as we believe this to be a subjective judgment.\footnote{A full list of author names, titles, genres and publishing years of the test set books can be found in Appendix A, Table \ref{app:test_set_books}.}

\paragraph{Training}
Firstly, we align the sentences of the English and Dutch versions of each book using Vecalign \cite{thompson-koehn-2019-vecalign}.
For the books in the test set, we manually discard sentences for which there existed no proper alignment, such as front matter sentences.
Additionally, we discard sentences with a cosine distance higher than 0.7 (2.3\% of all sentences).
Then, we normalise all punctuation using the MOSES toolkit.\footnote{http://www.statmt.org/moses/}
We then apply SentencePiece \cite{kudo-richardson-2018-sentencepiece} subword segmentation to the data. For this, we train a SentencePiece unigram model with a joint vocabulary for both languages and a vocabulary size of 32,000. 

We train a Transformer-based translation model using the Fairseq toolkit \cite{ott-etal-2019-fairseq}.
More specifically, we use the \textit{transformer\_iwslt\_de\_en} architecture. This is a Transformer base model with 6 encoder and decoder layers and an embedding dimension of 512. During training, we use an Adam optimiser, a learning rate of 5e-4, the loss function cross entropy with label smoothing 0.1 and the batch size is 64. Each model is trained until convergence with a patience of 3 epochs, using the BLEU score as a maximisation metric for finding the best checkpoint.

\paragraph{Decoding Strategies}
By default, we use beam search for decoding.
Reranking approaches rely heavily on the diversity of the translation hypotheses: if the hypotheses are all very similar, reranking them is not likely to have a large effect.
To ensure diverse hypotheses, we use a beam size of 20. Additionally, we experiment with decoding through diverse beam search \cite{vijayakumar2016diverse}. We follow \newcite{vijayakumar2016diverse} by using 3 groups, with a beam size of 21.
Beyond beam search, we investigate the effects of top-k and top-p sampling, with the default parameters and sampling size 10.

\subsection{Original-Text Classification}
\label{sec:cls}

\paragraph{Data}
We use a monolingual dataset of more than 7,000 Dutch books from varying original languages, authors and genres~\cite{toral-cranenburgh-nutters-2024}. %
For each book, we annotate whether it was originally written in Dutch.\footnote{The full annotation workflow can be found in Appendix C} We discard 2,182 books for which the original language is unclear or  that were not prose.
We make sure to avoid overlap with the parallel data set by removing any books that are also part of the parallel data.
Finally, we randomly sample 1,794 of the remaining 2,190 books as to match the total number of translated books, ensuring an equal distribution.
In total, we are left with over 3,500 books and over 29M sentences. We further divide these into data for system development and data for original-text classification. We use this data for reproducing previous work \cite{freitag-etal-2022-natural} and for training our classifier.
Additionally, we translate the classifier section of the monolingual data set using a reverse-direction trained version of the vanilla MT system (NL $\rightarrow$ EN), and then perform round-trip-translation (RTT) back to Dutch with the vanilla MT system, to obtain an MT version of the monolingual classifier data.
The full data size statistics and division in training, development and testing splits are listed in Table \ref{table:mono_data_stats}.

\begin{table}[t]
\begin{center}
\setlength{\tabcolsep}{3pt}
\resizebox{\columnwidth}{!}{
\begin{tabular}{llrrr}
\toprule
\multicolumn{5}{c}{\textbf{System development (90\%)}} \\
\bf Split &   \bf Orig.  & \bf \# Books & \bf \# Sentences & \bf \# Words \\
\midrule
Train (80\%) & Dutch & 1,291 & 8,576,756 & 10,425,656\\
  & Other & 1,291 & 12,470,149 & 165,263,466\\
 \midrule
  Dev (10\%)& Dutch & 162 & 1,005,832 & 12,533,406\\
    & Other & 162 & 1,546,057 & 19,723,706 \\
   \midrule
 Test (10\%) & Dutch & 162 & 1,189,690 & 14,721,914\\
     & Other & 162 & 1,573,499 & 20,968,346\\
\midrule
\multicolumn{5}{c}{\textbf{Original-text Classification (10\%)}} \\
\bf Split &   \bf Orig.  & \bf \# Books & \bf \# Sentences & \bf \# Words \\
\midrule
 Train (80\%) & Dutch & 143 & 982,114 & 11,528,789\\
  & Other & 143  &  139,0351 &  17,951,613 \\
  \midrule
  Test (20\%) & Dutch & 36 & 261,151 & 2,974,873 \\
    & Other & 36 &  340,950 & 4,283,604 \\
\bottomrule
\bf Total  &  &  3,588 & 29,336,549 & 376,130,733 \\ 
\bottomrule
\end{tabular}
}
\caption{Monolingual data set division and size.}
\label{table:mono_data_stats}
\end{center}
\end{table}

\paragraph{Training}

Currently, state-of-the-art performance for original-text detection is based on BERT \cite{devlin-etal-2019-bert}, as demonstrated by \newcite{pylypenko-etal-2021-comparing}.
We implement a similar system that distinguishes between original text and MT by training a binary classification model. We fine-tune Dutch language model BERTje \cite{devries2019bertje}. We train each model on the training split of the original-text classification data (see Table \ref{table:mono_data_stats}). 
We train models with batch size 128, accumulating gradients over 8 update steps, using the Adam optimiser~\cite{kingma-etal-2015-adam} with a learning rate of 3e-5. We use early stopping (patience 3) if validation performance does not improve.
On the held-out test set%
, the classifier achieves an accuracy of 85.9\%. It obtains a precision of 90.6\%, a recall of 80.2\% and the F1 score is 85.0\%.

\subsection{Baselines}

\paragraph{APE}
\newcite{freitag-etal-2019-ape} introduced Automatic Post-Editing (APE) as a post-hoc method to increase the `naturalness' of MT output. 
Following their approach, we train a post-processor that `translates' synthetic Dutch sequences into more natural Dutch sequences.
For training this system, we use the same data that was used to train the classifier (Section \ref{sec:cls}), consisting of RTT Dutch (which we use as source) and original Dutch (which we use as target). We train a model with the same architecture as the vanilla MT system.
We apply the post-processor to the output of the vanilla MT system, in an attempt to obtain a translation with a lexical diversity that is closer to HT.

\paragraph{Tagging}
Our second baseline is based on \newcite{freitag-etal-2022-natural}. We train an MT system that learns to differentiate between original and translated text during training.
This method requires both translated and original Dutch target samples.
The translated target samples are found in our parallel dataset. We use the same original Dutch samples that are used in training the translationese classifier.
Following \newcite{freitag-etal-2022-natural}, we then prepend ${<}orig{>}$ to the English source sentences that have original Dutch on the target side, and ${<}trans{>}$ for the source sentences that have translated Dutch. We train an MT system (same parameters as vanilla MT) on this data set. For inference, we prepend the source with ${<}orig{>}$, which prompts the model to produce a translation that exhibits characteristics that are often found in original Dutch. Note that, in contrast to APE, this method cannot be applied post-hoc.\footnote{Note that our implementation differs from \newcite{freitag-etal-2022-natural} in that they automatically differentiate natural and unnatural samples from a large parallel corpus using contrasting language models.}

\section{Evaluation}
\label{sec:eval}

We introduce three classes of metrics. Firstly, we look at general text metrics, which are commonly used for evaluating lexical diversity. Secondly, we use translation-specific metrics.
Lastly, we evaluate the general translation quality.

\subsection{General Text Metrics}
\noindent

\paragraph{TTR} The type-token ratio is the ratio of types (set of words) to tokens (actual words). A higher TTR indicates that more (different) words are used, which in turn indicates a higher lexical diversity. While this method is known to be influenced by the length of the text it is applied to, we report it because it is easy to interpret and widely used. %

\begin{table*}[t] 
    \centering
    \scalebox{0.87}{
    \begin{tabular}{l|ccc|ccc|cc}
    \toprule
        \textbf{Approach} & \textbf{TTR $\uparrow$}  &\textbf{Yule's I $\uparrow$} &  \textbf{MTLD $\uparrow$}  &  \textbf{PTF $\downarrow$} &  \textbf{CDU $\downarrow$} & \textbf{SynTTR $\uparrow$} &  \textbf{BLEU $\uparrow$} &  \textbf{COMET $\uparrow$} \\
    \midrule
        HT & 0.098 & 1.226 & 96.05 & 0.817 & 0.549 & 0.042 & - & - \\
         Vanilla MT & 0.089 & 0.951 & 90.21 & 0.832 & \textbf{0.550} & 0.040 & 32.32 &  0.824 \\
          APE & 0.092 & 0.985 &  90.59 & 0.827 & 0.554 & \textbf{0.041} & 30.39 & 0.808  \\
         Tagging & \textbf{0.095} & 1.111 & \textbf{94.08} & 0.829 & \textbf{0.550} & \textbf{0.041} & 31.33 & 0.807  \\
         \midrule
         Tailored RR \textit{(n=5)} & 0.091 & 1.002 & 92.46 & 0.829 & 0.552 & \textbf{0.041} & 30.92 & 0.815 \\
         Tailored RR \textit{(n=10)} & 0.091 & 1.013 & 93.26 & 0.829 & 0.547 & \textbf{0.041} & 30.07 & 0.810 \\
         Tailored RR \textit{(n=20)} & 0.092 & 1.010 & 93.27 & 0.830 & 0.558 & \textbf{0.041} & 28.98 & 0.802 \\
         \midrule
        Tailored RR \textit{(Top-k)} & \textbf{0.101} & \textbf{1.286} & 104.25 & \textbf{0.815} & 0.559 & \textbf{0.043} & 21.21 & 0.745\\
        Tailored RR \textit{(Top-p)} & 0.092 & 1.017 & 91.21 & 0.828 & 0.552 & \textbf{0.041} & 29.97 & 0.808\\
        Tailored RR \textit{(DBS)} & 0.092 & 1.010 & 92.70 & 0.828 & 0.553 & 0.040 & 29.36 &  0.805 \\
     \bottomrule
    \end{tabular}
    }
    \caption{Scores averaged across books, where RR stands for reranking. We provide results for multiple decoding strategies. Beam size is 20. Scores closest to HT are in bold font.}
    \label{tab:delta}
\end{table*}

\paragraph{Yule's I} As a metric that is less sensitive to variation in text length, we use Yule's I \cite{yule1944statistical}. We calculate this value as stated in Equation \ref{eq:yulesi}, where V is the size of the vocabulary (number of types) and \textit{t(i,N)} denotes the frequency of types which occur i times in a sample of length N.

\begin{equation}
	\text{Yule's}\hspace{0.33em} \text{I} = \frac{V^{2}}{\sum_{i=1} ^{V} \times t(i,N) - V}
	\label{eq:yulesi}
\end{equation}

\paragraph{MTLD} As an additional metric that has proven to be robust to document length variety, we use the measure of textual lexical diversity (MTLD), which is sequentially calculated as the `average length of sequential word strings in a text that maintain a given TTR value' \cite{mccarthy2005assessment}. %
We use the same TTR threshold (0.72) as \newcite{vanmassenhove-etal-2021-machine}.

We calculate these values using the \textit{LexicalRichness} Python library \cite{lex}.

\subsection{Translation-specific Metrics}
\noindent
\newcite{vanmassenhove-etal-2021-machine} introduce a novel automatic evaluation method for measuring lexical diversity in translations: Synonym Frequency Analysis (SFA).
It provides an insight into the diversity of lexical choices in translations.
For English words that have multiple translations in Dutch, it takes into account the frequency of these translation options.
We re-implement this method, as it was not implemented for our language pair before.
We first lemmatise each word in the source (English) side of our test set, using SpaCy (\textit{nl\_core\_news\_lg}).\footnote{https://spacy.io/models/nl\#nl\_core\_news\_lg} Next, we extract all possible translation options for the English adjectives, nouns and verbs by using a English-to-Dutch bilingual dictionary.\footnote{We use the dictionary from \url{https://freedict.org/downloads}. As an example, for the English adjective \emph{touching}, we find as Dutch translations: \textit{ontroerend}, \textit{aangrijpend}, \textit{emotioneel}, \textit{treffend}, \textit{roerend} and \textit{aandoenlijk}.}
Next, for each translation option, we count the number of occurrences in the MT output for each system.
The result is a vector which contains the occurrence frequency of each translation synonym for an English word. %

\begin{figure*}[t]
    \centering
\includegraphics[width=\textwidth]{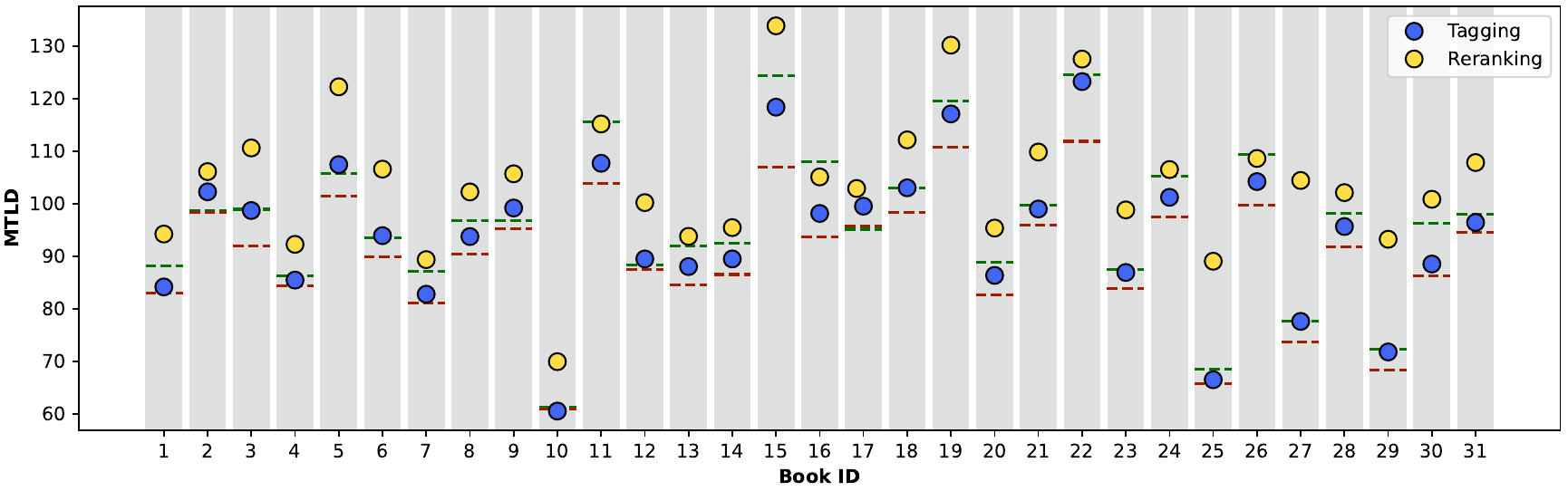}
    \caption{Per-book comparison of MTLD between the (rigid) tagging baseline and (tailored) reranking method, where green dotted lines are HT scores, and red dotted lines represent vanilla MT.}
    \label{fig:ttr}
\end{figure*}

\paragraph{PTF}
The primary translation frequency (PTF) is the average percentage (over all relevant source words) of times the most frequent translation option was chosen, from all translation options. The assumption is that if the output contains more secondary candidates, the text is more lexically diverse. 
We report the average PTF of all source words.

\paragraph{CDU}
The CDU is the cosine distance between the output vector for each source word and a vector of the same length with an equal distribution for each translation option (with the same total). We take the average CDU over all relevant source words to compute a final CDU.

\paragraph{SynTTR}
Lastly, we compute the SynTTR by dividing the number of types (the length of the set of all translation options) by the number of tokens (the sum of all translation options vectors).

\subsubsection{Translation Quality}
\noindent
We
also
calculate a general measure of translation quality, because the `naturalness' of a translation does not necessarily imply that a translation is a faithful representation of the source. A randomly generated string sequence might be very lexically diverse, but likely does not carry the source meaning.
Firstly, we calculate BLEU  \cite{papineni-etal-2002-bleu}, as implemented in SacreBLEU \cite{post-2018-call}. We use the default settings, which are case-sensitive.
Secondly, to account for the fact that BLEU does not necessarily evaluate meaning preservation, we additionally evaluate with COMET \cite{rei-etal-2020-comet}. English and Dutch are relatively high-resource languages, so we can use multilingual language embeddings. We report \textit{comet-score}, calculated with the default \textit{wmt22-comet-da}.
Still, it should be noted that these automatic metrics do not necessarily correlate strongly with human judgements, especially for literary translation.

\section{Results and Analysis}

\subsection{Quantitative Results}
We first discuss the results over all books.
Table \ref{tab:delta} shows the average results of measuring lexical diversity and general translation quality across the various approaches.
We find that vanilla MT indeed produces lexically poorer translations than HT, according to all our metrics.
While the scores of the APE baseline remain close to vanilla MT, our tailored reranking approach retrieves a lexical diversity that is closer to HT. This suggests that our method is a suitable alternative for post-hoc editing, given that one has access to the MT model for generating translation hypotheses.
The tagging baseline, which cannot be applied post-hoc, retrieves and MTLD and CDU that is on average closest to HT.
Importantly though, it should be noted that reranking and tagging are not mutually exclusive: one could apply reranking to the tagging baseline to increase or decrease lexical diversity further, where desired.
When we compare decoding strategies of the tailored reranking method, we first observe that using diverse beams search and choosing a larger $n$ retrieves at most slightly more diversity.
Especially top-k decoding retrieves a much higher lexical diversity.
However, tailored reranking comes with a compromise in terms of translation quality metrics.

Next, we demonstrate that these averages omit a more fine-grained view.
Figure \ref{fig:ttr} shows the difference in MTLD per book between vanilla MT, HT, tagging and our most diverse reranking system, based on top-k sampling, which is tailored to the LexDiv score of the original English book.\footnote{A similar figure with the posthoc baseline APE instead of tagging is shown in Appendix D.}
Our method renders almost every single book more lexically diverse than the tagging baseline.
In some cases, this makes the results closer to HT in terms of lexical diversity (e.g. 7, 13, 14, 16).
However, especially in cases where vanilla MT and HT are close already, this is not always true (e.g. 1, 3, 5).

\definecolor{mintbg}{rgb}{.63,.79,.95}
\definecolor{corn}{rgb}{0.98, 0.93, 0.36}
\definecolor{fig1yellow}{rgb}{1, 0.95, 0.8}

\begin{table*}[t]
    \centering
    \scalebox{0.95}{
    \begin{tabular}{l|l|l}
    \toprule
        \bf{Ex. \#} &\bf{Approach} & \bf{Text} \\
    \midrule
        1 & Source & The \colorbox{corn}{kid} had no mother.\\
         &HT & Dat \colorbox{corn}{joch} heeft geen moeder gehad. \\
         &Vanilla MT & Het \colorbox{corn}{kind} had geen moeder. \\
         &Tagging & De \colorbox{corn}{jongen} had geen moeder. \\
        & Tailored RR & Het \colorbox{corn}{joch} had geen moeder. \\
    \midrule
        2 & Source & He \colorbox{corn}{shipped} his oars and brought a small \colorbox{mintbg}{line} from under the bow.\\
         &HT & Hij \colorbox{corn}{haalde} de riemen \colorbox{corn}{in} en pakte een kleine \colorbox{mintbg}{lijn} die voor in de boot lag. \\
         &Vanilla MT & Hij \colorbox{corn}{trok} zijn riemen \colorbox{corn}{aan} en haalde een klein \colorbox{mintbg}{lijntje} onder de boeg vandaan. \\
         &Tagging & Hij \colorbox{corn}{verscheurde} zijn riemen en haalde een klein \colorbox{mintbg}{streepje} onder de boeg vandaan. \\
        & Tailored RR & Hij \colorbox{corn}{haalde} zijn riemen en trok er een kleine \colorbox{mintbg}{lijn} voor onder de boot vandaan. \\
    \midrule
        3 & Source & In long \colorbox{corn}{shaky} strokes Sargent \colorbox{mintbg}{copied} the data. \\
         &HT & In lange \colorbox{corn}{beverige} halen  \colorbox{mintbg}{kopieerde} Sargent de gegevenheden. \\
         &Vanilla MT & Met lange, \colorbox{corn}{bevende} slagen  \colorbox{mintbg}{kopieerde} Sargent de gegevens. \\
         &Tagging & Met lange \colorbox{corn}{bevende} halen  \colorbox{mintbg}{kopieerde} Sargent de gegevens. \\
        & Tailored RR & Met lange \colorbox{corn}{beverige} halen  \colorbox{mintbg}{schreef} Sargent de data  \colorbox{mintbg}{over}. \\
    \bottomrule
    \end{tabular}
    }
    \caption{Examples to highlight surface-level differences between the systems' output translations, where Tailored RR uses top-k sampling.}
    \label{tab:surface-level}
\end{table*}

\subsection{Surface-level Inspection}
The output translations were inspected by a native speaker. Table 4 shows three examples of how translations differ between vanilla MT, tagging and tailored reranking (with top-k sampling).
In Example 1 (from book 1, \textit{Sunset Park}), we see that the English noun `kid' is translated as \textit{joch} (`boy') in the human translation, which is less common than the vanilla MT's \textit{kind} (`child') and tagging's \textit{jongen} (`boy'). This is recovered by our tailored reranking system, which uses \textit{joch} too. %

Example 2 is taken from book 10, \textit{The Old Man and the Sea}, which has low lexical diversity by default (see Section \ref{sec:why}). This is not taken into account by the tagging baseline: the English `shipped' is translated as a less common (and wrong) \textit{verscheurde} (`shredded'). 
The tailored reranking system (\textit{haalde}, `brought') is closest to HT (\textit{haalde in}, `brought in').
Additionally, the tagging baseline wrongly translates the English `line' as \textit{streepje} (`small stripe'), while  tailored reranking (\textit{lijn}, `line') is again identical to HT. This case illustrates that choosing a more common translation synonym, which may for instance results in a lower PTF, may for some books be closer to HT.

By contrast, in Example 3 from the more lexically diverse \textit{Ulysses} (book 15), the tagging baseline stays closer to vanilla MT: both translate `shaky' as \textit{bevend} (`trembling').
Tailored reranking outputs \textit{beverig} (`shaky'), which is again recovering the HT. 
Furthermore, tailored reranking deviates from all other systems (and HT) by translating `copied' into the translation synonym \textit{schreef over} (copying something by writing). This case may illustrate why the tailored reranking based on top-k sampling surpasses the other systems in the overall metrics.

\subsection{Ranks and Lexical Diversity}

\begin{figure}[t]
    \centering
    \includegraphics[width=0.45\textwidth]{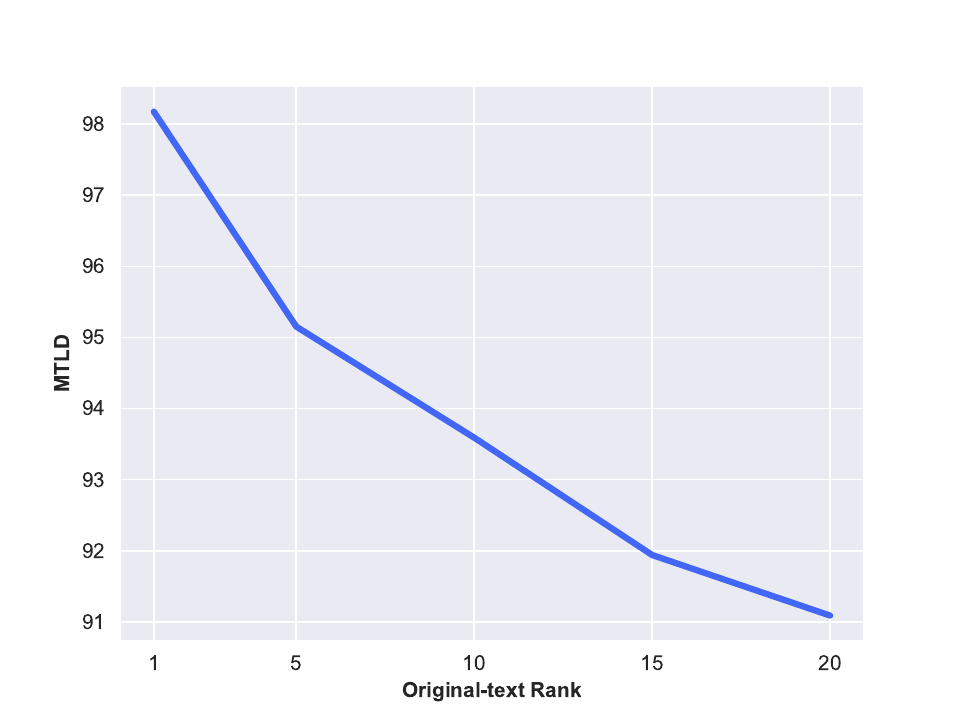}
    \caption{Change in MTLD for choosing different ranks, where beam size is 20 and $n=20$.}
    \label{fig:flex-demo}
\end{figure}

\begin{figure*}[h]
    \centering
    \includegraphics[width=\textwidth]{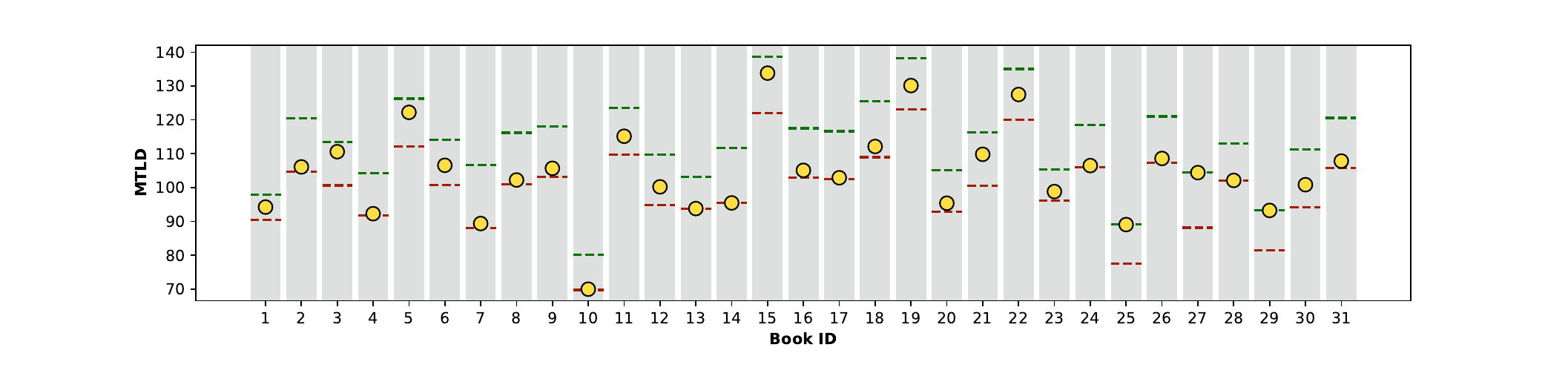}
    \caption{MTLD for highest (green), lowest (red) and tailored (yellow) original-text rank.}
    \label{fig:tailoring}
\end{figure*}

So far, we have assumed that reranking based on the probability of a candidate being original text leads to more lexically diverse output translations. Here, we verify whether choosing a lower probability of a candidate being original, actually implies lexically poorer output translations (Figure \ref{fig:flex-demo}).
For the vanilla MT system with beam size 20 and $n=20$, we first calculate the original-text probability for each translation hypothesis.
Similar to reranking, we sort the hypotheses according to this probability. Then, instead of binning, we choose the $n^{th}$ rank, and calculate lexical diversity of the output.
Figure \ref{fig:flex-demo} shows the change in MTLD scores for choosing a lower diversity rank. We observe that indeed, choosing a lower rank retrieves lower diversity (note that there, a higher rank represents a smaller original-text probability). This trend holds for TTR and Yule's I as well (see Appendix E).

\subsection{Tailoring and Lexical Diversity}
To further demonstrate the effect of a \textit{tailored} approach in lexical diversity, we compare MTLD scores of a top-k reranking system that always outputs the highest original-text probability, with the same system that always outputs the lowest, and a tailored version. Figure \ref{fig:tailoring} shows the results.
Firstly, we observe that, in every case, choosing a rank that represents lower original-text probability retrieves a lower MTLD score than choosing the opposite. This corroborates the findings from the previous section.
Next, we look into how the tailored reranking affects the output lexical diversity.
In Section \ref{sec:why}, we used \textit{The Old Man and the Sea} (book 10) as an example of a book with a low default lexical diversity. We observe that our tailored reranking system outputs the lowest original-text probability rank for this book, resulting in a lower MTLD score.
For the example from Section \ref{sec:why} of a lexically rich book, \textit{Ulysses} (book 15), our tailored system outputs a rank with a original-text probability higher than the minimum, thus retrieving an MTLD score that is higher.
This shows that tailoring is at least somewhat intuitive.

\section{Conclusion}

We have argued for flexible recovery of lexical diversity in literary MT.
We showed that default diversity varies per book in our dataset, and that this lexical diversity is partially lost through MT.
We presented the first approach towards tailored rescoring of translation candidates, which matches HT more closely than previous baselines for some books.
Future work could explore how our method can be combined with previous work, as it is in principle model-agnostic.
Investigations with document-level translation, instead of sentence-level translation only, could provide additional insights.
Furthermore, it may be useful to address this task at an even finer-grained level, by exploring diversity reranking on a sequence-level, instead of a book-level.

\section*{Limitations}
In this paper, we addressed the increase of lexical diversity in literary MT. However, it should be noted that this is does not encompass writing style as a whole.
We evaluated our approach on one high-resource language pair that consist of relatively similar languages, in one translation direction. For the domain of literary translation, we find this to be difficult to avoid. Still, experiments with more languages and resource-scenarios may retrieve interesting results.
Moreover, while our data is transparent in the sense that we know and can explain exactly what it contains, we cannot distribute the data ourselves because of copyright.
Lastly, we acknowledge that large-scale human evaluation could give useful insights into the differences between the systems.

\section*{Acknowledgements}
This work was supported by a \textit{Semper Ardens: Accelerate research grant (CF21-0454)} from the Carlsberg Foundation.
We thank the Center for Information Technology of the University of Groningen for their support and for providing access to the Peregrine and Hábrók high performance computing cluster.

\bibliography{custom.bib}

\begin{thebibliography}{}

\bibitem[\protect\citename{Arcadinho \bgroup et al.\egroup }2022]{arcadinho-etal-2022-t5ql}
Arcadinho, Samuel~David, David Aparicio, Hugo Veiga, and Antonio Alegria.
\newblock 2022.
\newblock {T}5{QL}: Taming language models for {SQL} generation.
\newblock In Bosselut, Antoine, Khyathi Chandu, Kaustubh Dhole, Varun Gangal, Sebastian Gehrmann, Yacine Jernite, Jekaterina Novikova, and Laura Perez-Beltrachini, editors, {\em Proceedings of the 2nd Workshop on Natural Language Generation, Evaluation, and Metrics (GEM)}, pages 276--286, Abu Dhabi, United Arab Emirates (Hybrid), December. Association for Computational Linguistics.

\bibitem[\protect\citename{Baker}1993]{baker1993corpus}
Baker, Mona.
\newblock 1993.
\newblock Corpus linguistics and translation studies—implications and applications.
\newblock In {\em Text and Technology}, page 233. John Benjamins.

\bibitem[\protect\citename{Baroni and Bernardini}2005]{baroni2006}
Baroni, Marco and Silvia Bernardini.
\newblock 2005.
\newblock {A New Approach to the Study of Translationese: Machine-learning the Difference between Original and Translated Text}.
\newblock {\em Literary and Linguistic Computing}, 21(3):259--274, 08.

\bibitem[\protect\citename{Boase-Beier}2011]{boase2011critical}
Boase-Beier, Jean.
\newblock 2011.
\newblock {\em A critical introduction to translation studies}.
\newblock Bloomsbury Publishing.

\bibitem[\protect\citename{Collins and Koo}2005]{collins2005discriminative}
Collins, Michael and Terry Koo.
\newblock 2005.
\newblock Discriminative reranking for natural language parsing.
\newblock {\em Computational Linguistics}, 31(1):25--70.

\bibitem[\protect\citename{de Vries \bgroup et al.\egroup }2019]{devries2019bertje}
de~Vries, Wietse, Andreas van Cranenburgh, Arianna Bisazza, Tommaso Caselli, Gertjan~van Noord, and Malvina Nissim.
\newblock 2019.
\newblock {BERTje}: {A} {Dutch} {BERT} {Model}.
\newblock arXiv:1912.09582, December.

\bibitem[\protect\citename{Delabastita}2011]{delabastita2011literary}
Delabastita, Dirk.
\newblock 2011.
\newblock Literary translation.
\newblock {\em Handbook of translation studies}, 2:69--78.

\bibitem[\protect\citename{Devlin \bgroup et al.\egroup }2019]{devlin-etal-2019-bert}
Devlin, Jacob, Ming-Wei Chang, Kenton Lee, and Kristina Toutanova.
\newblock 2019.
\newblock {BERT}: Pre-training of deep bidirectional transformers for language understanding.
\newblock In {\em Proceedings of the 2019 Conference of the North {A}merican Chapter of the Association for Computational Linguistics: Human Language Technologies, Volume 1 (Long and Short Papers)}, pages 4171--4186, Minneapolis, Minnesota, June. Association for Computational Linguistics.

\bibitem[\protect\citename{Dutta~Chowdhury \bgroup et al.\egroup }2022]{dutta-chowdhury-etal-2022-towards}
Dutta~Chowdhury, Koel, Rricha Jalota, Cristina Espa{\~n}a-Bonet, and Josef Genabith.
\newblock 2022.
\newblock Towards debiasing translation artifacts.
\newblock In {\em Proceedings of the 2022 Conference of the North American Chapter of the Association for Computational Linguistics: Human Language Technologies}, pages 3983--3991, Seattle, United States, July. Association for Computational Linguistics.

\bibitem[\protect\citename{Fischer and L{\"a}ubli}2020]{fischer-laubli-2020-whats}
Fischer, Lukas and Samuel L{\"a}ubli.
\newblock 2020.
\newblock What{'}s the difference between professional human and machine translation? a blind multi-language study on domain-specific {MT}.
\newblock In {\em Proceedings of the 22nd Annual Conference of the European Association for Machine Translation}, pages 215--224, Lisboa, Portugal, November. European Association for Machine Translation.

\bibitem[\protect\citename{Freitag \bgroup et al.\egroup }2019]{freitag-etal-2019-ape}
Freitag, Markus, Isaac Caswell, and Scott Roy.
\newblock 2019.
\newblock {APE} at scale and its implications on {MT} evaluation biases.
\newblock In {\em Proceedings of the Fourth Conference on Machine Translation (Volume 1: Research Papers)}, pages 34--44, Florence, Italy, August. Association for Computational Linguistics.

\bibitem[\protect\citename{Freitag \bgroup et al.\egroup }2022]{freitag-etal-2022-natural}
Freitag, Markus, David Vilar, David Grangier, Colin Cherry, and George Foster.
\newblock 2022.
\newblock A natural diet: Towards improving naturalness of machine translation output.
\newblock In {\em Findings of the Association for Computational Linguistics: ACL 2022}, pages 3340--3353, Dublin, Ireland, May. Association for Computational Linguistics.

\bibitem[\protect\citename{Heaton}1970]{heaton1970style}
Heaton, CP.
\newblock 1970.
\newblock Style in the old man and the sea.
\newblock {\em Style}, pages 11--27.

\bibitem[\protect\citename{Jalota \bgroup et al.\egroup }2023]{jalota-etal-2023-translating}
Jalota, Rricha, Koel Chowdhury, Cristina Espa{\~n}a-Bonet, and Josef van Genabith.
\newblock 2023.
\newblock Translating away translationese without parallel data.
\newblock In Bouamor, Houda, Juan Pino, and Kalika Bali, editors, {\em Proceedings of the 2023 Conference on Empirical Methods in Natural Language Processing}, pages 7086--7100, Singapore, December. Association for Computational Linguistics.

\bibitem[\protect\citename{Jimenez-Crespo}2023]{jimenez-crespo-2023-translationese}
Jimenez-Crespo, Miguel~A.
\newblock 2023.
\newblock {``}translationese{''} (and {``}post-editese{''}?) no more: on importing fuzzy conceptual tools from translation studies in {MT} research.
\newblock In Nurminen, Mary, Judith Brenner, Maarit Koponen, Sirkku Latomaa, Mikhail Mikhailov, Frederike Schierl, Tharindu Ranasinghe, Eva Vanmassenhove, Sergi~Alvarez Vidal, Nora Aranberri, Mara Nunziatini, Carla~Parra Escart{\'\i}n, Mikel Forcada, Maja Popovic, Carolina Scarton, and Helena Moniz, editors, {\em Proceedings of the 24th Annual Conference of the European Association for Machine Translation}, pages 261--268, Tampere, Finland, June. European Association for Machine Translation.

\bibitem[\protect\citename{Kingma and Ba}2015]{kingma-etal-2015-adam}
Kingma, Diederik~P. and Jimmy Ba.
\newblock 2015.
\newblock Adam: A method for stochastic optimization.
\newblock In Bengio, Yoshua and Yann LeCun, editors, {\em 3rd International Conference on Learning Representations (ICLR 2015)}.

\bibitem[\protect\citename{Koppel and Ordan}2011]{koppel-ordan-2011-translationese}
Koppel, Moshe and Noam Ordan.
\newblock 2011.
\newblock Translationese and its dialects.
\newblock In {\em Proceedings of the 49th Annual Meeting of the Association for Computational Linguistics: Human Language Technologies}, pages 1318--1326, Portland, Oregon, USA, June. Association for Computational Linguistics.

\bibitem[\protect\citename{Kudo and Richardson}2018]{kudo-richardson-2018-sentencepiece}
Kudo, Taku and John Richardson.
\newblock 2018.
\newblock {S}entence{P}iece: A simple and language independent subword tokenizer and detokenizer for neural text processing.
\newblock In {\em Proceedings of the 2018 Conference on Empirical Methods in Natural Language Processing: System Demonstrations}, pages 66--71, Brussels, Belgium, November. Association for Computational Linguistics.

\bibitem[\protect\citename{Lakew \bgroup et al.\egroup }2018]{lakew-etal-2018-comparison}
Lakew, Surafel~Melaku, Mauro Cettolo, and Marcello Federico.
\newblock 2018.
\newblock A comparison of transformer and recurrent neural networks on multilingual neural machine translation.
\newblock In {\em Proceedings of the 27th International Conference on Computational Linguistics}, pages 641--652, Santa Fe, New Mexico, USA, August. Association for Computational Linguistics.

\bibitem[\protect\citename{Lee \bgroup et al.\egroup }2021]{lee-etal-2021-discriminative}
Lee, Ann, Michael Auli, and Marc{'}Aurelio Ranzato.
\newblock 2021.
\newblock Discriminative reranking for neural machine translation.
\newblock In Zong, Chengqing, Fei Xia, Wenjie Li, and Roberto Navigli, editors, {\em Proceedings of the 59th Annual Meeting of the Association for Computational Linguistics and the 11th International Joint Conference on Natural Language Processing (Volume 1: Long Papers)}, pages 7250--7264, Online, August. Association for Computational Linguistics.

\bibitem[\protect\citename{Liu and Liu}2021]{liu-liu-2021-simcls}
Liu, Yixin and Pengfei Liu.
\newblock 2021.
\newblock {S}im{CLS}: A simple framework for contrastive learning of abstractive summarization.
\newblock In Zong, Chengqing, Fei Xia, Wenjie Li, and Roberto Navigli, editors, {\em Proceedings of the 59th Annual Meeting of the Association for Computational Linguistics and the 11th International Joint Conference on Natural Language Processing (Volume 2: Short Papers)}, pages 1065--1072, Online, August. Association for Computational Linguistics.

\bibitem[\protect\citename{Matusov}2019]{matusov-2019-challenges}
Matusov, Evgeny.
\newblock 2019.
\newblock The challenges of using neural machine translation for literature.
\newblock In Hadley, James, Maja Popovi{\'c}, Haithem Afli, and Andy Way, editors, {\em Proceedings of the Qualities of Literary Machine Translation}, pages 10--19, Dublin, Ireland, August. European Association for Machine Translation.

\bibitem[\protect\citename{McCarthy}2005]{mccarthy2005assessment}
McCarthy, Philip~M.
\newblock 2005.
\newblock {\em An assessment of the range and usefulness of lexical diversity measures and the potential of the measure of textual, lexical diversity (MTLD)}.
\newblock {Ph.D.} thesis, The University of Memphis.

\bibitem[\protect\citename{Ott \bgroup et al.\egroup }2019]{ott-etal-2019-fairseq}
Ott, Myle, Sergey Edunov, Alexei Baevski, Angela Fan, Sam Gross, Nathan Ng, David Grangier, and Michael Auli.
\newblock 2019.
\newblock fairseq: A fast, extensible toolkit for sequence modeling.
\newblock In {\em Proceedings of the 2019 Conference of the North {A}merican Chapter of the Association for Computational Linguistics (Demonstrations)}, pages 48--53, Minneapolis, Minnesota, June. Association for Computational Linguistics.

\bibitem[\protect\citename{Papineni \bgroup et al.\egroup }2002]{papineni-etal-2002-bleu}
Papineni, Kishore, Salim Roukos, Todd Ward, and Wei-Jing Zhu.
\newblock 2002.
\newblock {B}leu: a method for automatic evaluation of machine translation.
\newblock In {\em Proceedings of the 40th Annual Meeting of the Association for Computational Linguistics}, pages 311--318, Philadelphia, Pennsylvania, USA, July. Association for Computational Linguistics.

\bibitem[\protect\citename{Popel \bgroup et al.\egroup }2020]{popel2020transforming}
Popel, Martin, Marketa Tomkova, Jakub Tomek, {\L}ukasz Kaiser, Jakob Uszkoreit, Ond{\v{r}}ej Bojar, and Zden{\v{e}}k {\v{Z}}abokrtsk{\`y}.
\newblock 2020.
\newblock Transforming machine translation: a deep learning system reaches news translation quality comparable to human professionals.
\newblock {\em Nature communications}, 11(1):1--15.

\bibitem[\protect\citename{Post}2018]{post-2018-call}
Post, Matt.
\newblock 2018.
\newblock A call for clarity in reporting {BLEU} scores.
\newblock In {\em Proceedings of the Third Conference on Machine Translation: Research Papers}, pages 186--191, Brussels, Belgium, October. Association for Computational Linguistics.

\bibitem[\protect\citename{Pylypenko \bgroup et al.\egroup }2021]{pylypenko-etal-2021-comparing}
Pylypenko, Daria, Kwabena Amponsah-Kaakyire, Koel Dutta~Chowdhury, Josef van Genabith, and Cristina Espa{\~n}a-Bonet.
\newblock 2021.
\newblock Comparing feature-engineering and feature-learning approaches for multilingual translationese classification.
\newblock In {\em Proceedings of the 2021 Conference on Empirical Methods in Natural Language Processing}, pages 8596--8611, Online and Punta Cana, Dominican Republic, November. Association for Computational Linguistics.

\bibitem[\protect\citename{Rabinovich and Wintner}2015]{rabinovich-wintner-2015-unsupervised}
Rabinovich, Ella and Shuly Wintner.
\newblock 2015.
\newblock Unsupervised identification of translationese.
\newblock {\em Transactions of the Association for Computational Linguistics}, 3:419--432.

\bibitem[\protect\citename{Rei \bgroup et al.\egroup }2020]{rei-etal-2020-comet}
Rei, Ricardo, Craig Stewart, Ana~C Farinha, and Alon Lavie.
\newblock 2020.
\newblock {COMET}: A neural framework for {MT} evaluation.
\newblock In {\em Proceedings of the 2020 Conference on Empirical Methods in Natural Language Processing (EMNLP)}, pages 2685--2702, Online, November. Association for Computational Linguistics.

\bibitem[\protect\citename{Riera}2022]{jorge-braga-riera-2022}
Riera, Jorge~Braga.
\newblock 2022.
\newblock Literatura-traducción.
\newblock {\em Enciclopedia de Traducción e Interpretación}.

\bibitem[\protect\citename{Shen \bgroup et al.\egroup }2004]{shen2004discriminative}
Shen, Libin, Anoop Sarkar, and Franz~Josef Och.
\newblock 2004.
\newblock Discriminative reranking for machine translation.
\newblock In {\em Proceedings of the Human Language Technology Conference of the North American Chapter of the Association for Computational Linguistics: HLT-NAACL 2004}, pages 177--184.

\bibitem[\protect\citename{Shen}2022]{lex}
Shen, Lucas.
\newblock 2022.
\newblock {LexicalRichness: A small module to compute textual lexical richness}.

\bibitem[\protect\citename{Thompson and Koehn}2019]{thompson-koehn-2019-vecalign}
Thompson, Brian and Philipp Koehn.
\newblock 2019.
\newblock {V}ecalign: Improved sentence alignment in linear time and space.
\newblock In {\em Proceedings of the 2019 Conference on Empirical Methods in Natural Language Processing and the 9th International Joint Conference on Natural Language Processing (EMNLP-IJCNLP)}, pages 1342--1348, Hong Kong, China, November. Association for Computational Linguistics.

\bibitem[\protect\citename{Toral and Way}2015]{toral2015machine}
Toral, Antonio and Andy Way.
\newblock 2015.
\newblock Machine-assisted translation of literary text: A case study.
\newblock {\em Translation Spaces}, 4(2):240--267.

\bibitem[\protect\citename{Toral \bgroup et al.\egroup }2018]{toral-etal-2018-attaining}
Toral, Antonio, Sheila Castilho, Ke~Hu, and Andy Way.
\newblock 2018.
\newblock Attaining the unattainable? {R}eassessing claims of human parity in neural machine translation.
\newblock In {\em Proceedings of the Third Conference on Machine Translation: Research Papers}, pages 113--123, Brussels, Belgium, October. Association for Computational Linguistics.

\bibitem[\protect\citename{Toral \bgroup et al.\egroup }2024]{toral-cranenburgh-nutters-2024}
Toral, Antonio, {Andreas van} Cranenburgh, and Tia Nutters, 2024.
\newblock {\em Literary-adapted machine translation in a well-resourced language pair: Explorations with More Data and Wider Contexts}, pages 27--52.
\newblock Routledge.

\bibitem[\protect\citename{Trotta}2014]{trotta2014creativity}
Trotta, Joe.
\newblock 2014.
\newblock Creativity, playfulness and linguistic carnivalization in james joyce's ulysses.

\bibitem[\protect\citename{van~der Werff \bgroup et al.\egroup }2022]{van-der-werff-etal-2022-automatic}
van~der Werff, Tobias, Rik van Noord, and Antonio Toral.
\newblock 2022.
\newblock Automatic discrimination of human and neural machine translation: A study with multiple pre-trained models and longer context.
\newblock In {\em Proceedings of the 23rd Annual Conference of the European Association for Machine Translation}, pages 161--170, Ghent, Belgium, June. European Association for Machine Translation.

\bibitem[\protect\citename{Vanmassenhove \bgroup et al.\egroup }2019]{vanmassenhove2019lost}
Vanmassenhove, Eva, Dimitar Shterionov, and Andy Way.
\newblock 2019.
\newblock Lost in translation: Loss and decay of linguistic richness in machine translation.
\newblock In {\em Proceedings of Machine Translation Summit XVII: Research Track}, pages 222--232.

\bibitem[\protect\citename{Vanmassenhove \bgroup et al.\egroup }2021]{vanmassenhove-etal-2021-machine}
Vanmassenhove, Eva, Dimitar Shterionov, and Matthew Gwilliam.
\newblock 2021.
\newblock Machine translationese: Effects of algorithmic bias on linguistic complexity in machine translation.
\newblock In {\em Proceedings of the 16th Conference of the European Chapter of the Association for Computational Linguistics: Main Volume}, pages 2203--2213, Online, April. Association for Computational Linguistics.

\bibitem[\protect\citename{Vaswani \bgroup et al.\egroup }2017]{vaswani2017attention}
Vaswani, Ashish, Noam Shazeer, Niki Parmar, Jakob Uszkoreit, Llion Jones, Aidan~N Gomez, {\L}ukasz Kaiser, and Illia Polosukhin.
\newblock 2017.
\newblock Attention is all you need.
\newblock {\em Advances in neural information processing systems}, 30.

\bibitem[\protect\citename{Vijayakumar \bgroup et al.\egroup }2016]{vijayakumar2016diverse}
Vijayakumar, Ashwin~K, Michael Cogswell, Ramprasath~R Selvaraju, Qing Sun, Stefan Lee, David Crandall, and Dhruv Batra.
\newblock 2016.
\newblock Diverse beam search: Decoding diverse solutions from neural sequence models.
\newblock {\em arXiv preprint arXiv:1610.02424}.

\bibitem[\protect\citename{Volansky \bgroup et al.\egroup }2015]{volansky2015features}
Volansky, Vered, Noam Ordan, and Shuly Wintner.
\newblock 2015.
\newblock On the features of translationese.
\newblock {\em Digital Scholarship in the Humanities}, 30(1):98--118.

\bibitem[\protect\citename{Wright}2016]{wright2016literary}
Wright, Chantal.
\newblock 2016.
\newblock {\em Literary translation}.
\newblock Routledge.

\bibitem[\protect\citename{Yin and Neubig}2019]{yin-neubig-2019-reranking}
Yin, Pengcheng and Graham Neubig.
\newblock 2019.
\newblock Reranking for neural semantic parsing.
\newblock In Korhonen, Anna, David Traum, and Llu{\'\i}s M{\`a}rquez, editors, {\em Proceedings of the 57th Annual Meeting of the Association for Computational Linguistics}, pages 4553--4559, Florence, Italy, July. Association for Computational Linguistics.

\bibitem[\protect\citename{Yule}1944]{yule1944statistical}
Yule, C~Udny.
\newblock 1944.
\newblock {\em The statistical study of literary vocabulary}.
\newblock Cambridge University Press.

\end{thebibliography}
\bibliographystyle{eamt24}

\onecolumn
\newpage

\appendix{{\bf A \hspace{1em} Test set novels}}
\label{app:test_set_books}

\begin{table}[h!]
\begin{center}
\scalebox{0.78}{
\begin{tabular}{rllrl}
\toprule
\bf ID &\bf Author &  \bf Title & \bf Year Published & \bf Genre \\
\midrule
1&Paul Auster & Sunset Park  & 2010 & Literary fiction \\
2&David Baldacci & Divine Justice & 2008 & Thriller, suspense\\
3&Julian Barnes & The Sense of an Ending & 2011 & Literary fiction \\ 
4&John Boyne & The Boy in the Striped Pyjamas & 2006 & Historical fiction \\ 
5&John le Carré & Our Kind of Traitor & 2010 & Thriller, spy fiction \\ 
6&Jonathan Franzen & The Corrections & 2001 & Literary fiction \\ 
7&Nicci French & Blue Monday: A Frieda Klein Mystery & 2011 & Thriller, suspense \\ 
8&William Golding & Lord of the Flies & 1954 & Literary fiction \\ 
9&John Grisham & The Confession & 2010 & Thriller, suspense \\ 
10&Ernest Hemingway & The Old Man and the Sea & 1952 & Literary fiction \\ 
11&Patricia Highsmith & Ripley Under Water & 1991 & Thriller, suspense \\ 
12&Khaled Hosseini & A Thousand Splendid Suns & 2007 & Literary fiction \\ 
13&John Irving & Last Night in Twisted River & 2009 & Literary fiction\\ 
14&E.L. James & Fifty Shades of Grey & 2011 & Erotic thriller \\ 
15&James Joyce & Ulysses & 1922 & Literary fiction \\ 
16&Jack Kerouac & On the Road & 1957 & Literary fiction \\ 
17&Stephen King & 11/22/63 & 2011 & Science-fiction \\ 
18&Sophie Kinsella & Shopaholic and Baby & 2007 & Popular literature \\ 
19&David Mitchell & The Thousand Autumns of Jacob de Zoet & 2010 & Historical fiction \\ 
20&George Orwell & 1984 & 1949 & Literary fiction\\ 
21&James Patterson & The Quickie & 2007 & Thriller, suspense \\ 
22&Thomas Pynchon & Gravity's Rainbow & 1973 & Historical fiction\\ 
23&Philip Roth & The Plot Against America & 2004 & Political fiction \\ 
24&J.K. Rowling & Harry Potter and the Deathly Hallows & 2007 & Fantasy \\ 
25&J.D. Salinger & The Catcher in the Rye & 1951 & Literary fiction \\ 
26&Karin Slaughter & Fractured & 2008 & Thriller, suspense \\ 
27&John Steinbeck & The Grapes of Wrath & 1939 & Literary fiction \\ 
28&J.R.R Tolkien & The Return of the King & 1955 & Fantasy \\ 
29&Mark Twain & Adventures of Huckleberry Finn & 1884 & Literary fiction \\ 
30&Oscar Wilde & The Picture of Dorian Gray & 1890 & Literary fiction \\ 
31&Irvin D. Yalom & The Spinoza Problem & 2012 & Historical fiction \\ 
\bottomrule
\end{tabular}
}
\caption{Information on test set books.}
\label{table:test_set_info}
\end{center}
\end{table}

\vspace{5em}

\appendix{{\bf B \hspace{1em} Regression plots for human translation vs. original text lexical diversity}}
\label{app:regr_plots}

\begin{figure*}[h]
    \includegraphics[width=0.33\textwidth]{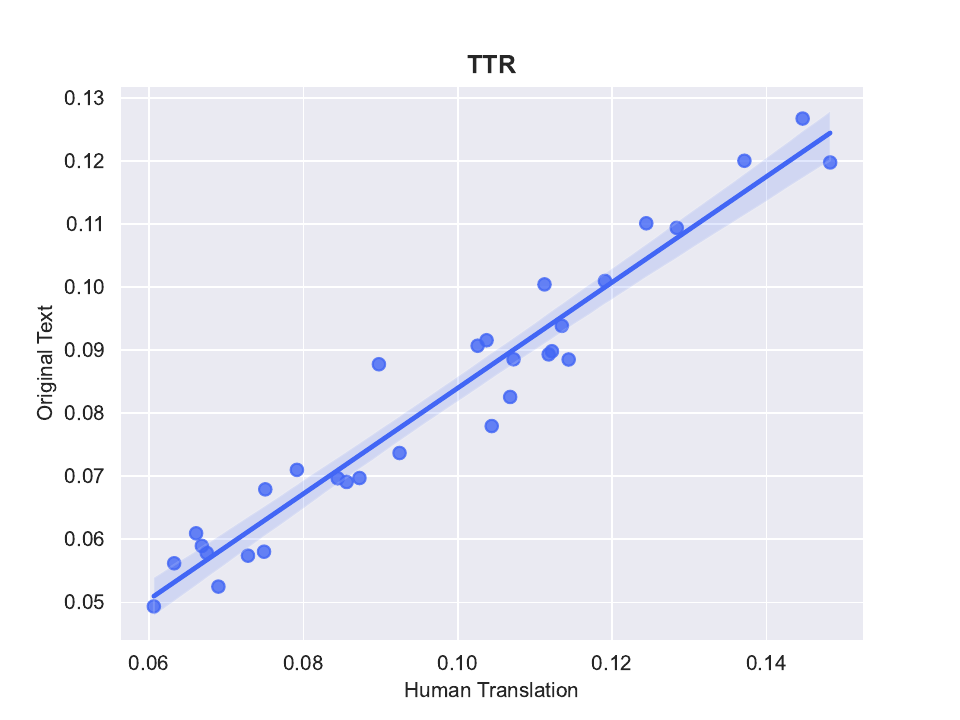}
    \includegraphics[width=0.33\textwidth]{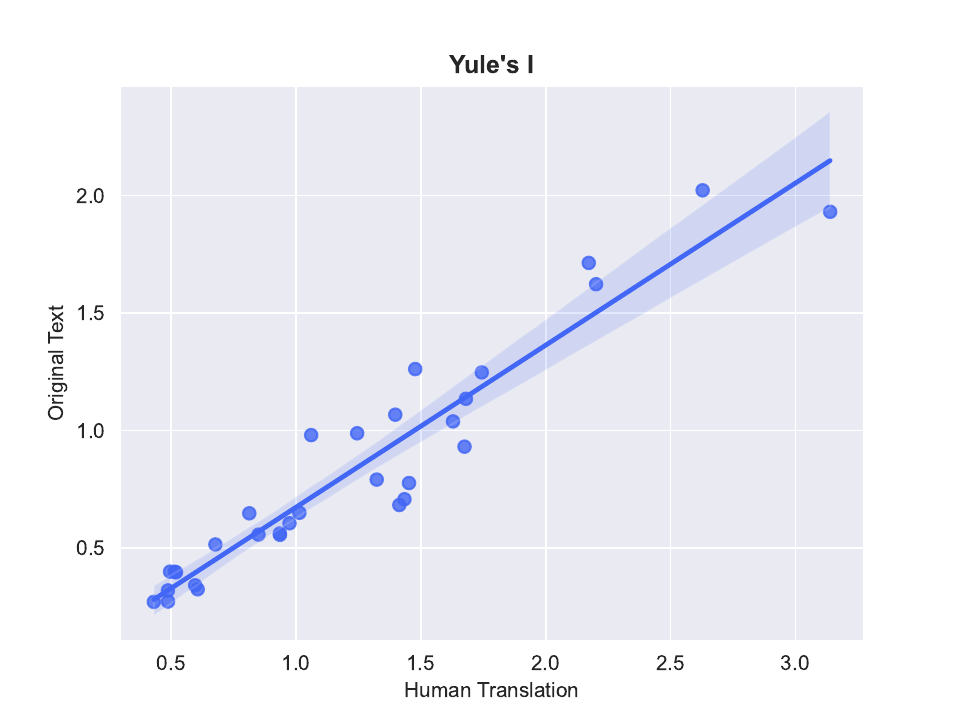}
    \includegraphics[width=0.33\textwidth]{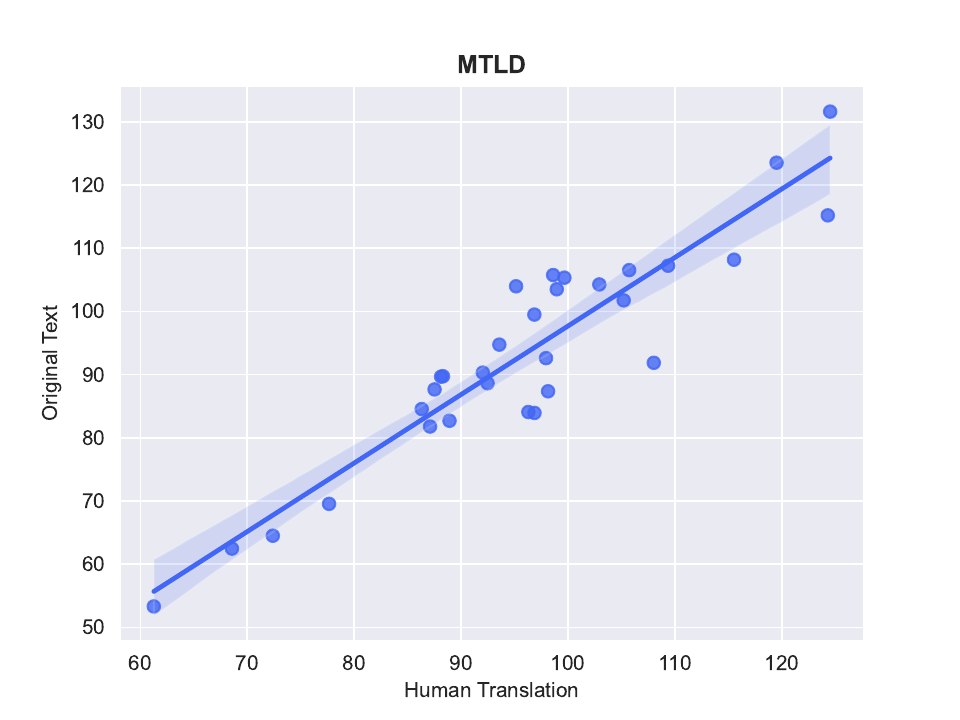}
    \caption{Regression plots for TTR, Yule's I and MTLD, with on the y-axis the scores for the original (English) versions, and on the x-axis those for human translations.}
\end{figure*}

\newpage

\appendix{{\bf C \hspace{1em} Annotation workflow for monolingual Dutch books}}
\label{app:ann_workflow}

\begin{enumerate}
    \item Check whether the book is prose: we generally discard other forms of literature such as poetry and plays and annotate this in category 3 (no label).
    \item Check whether the original language of the book is listed on the website of the National Dutch Library.\footnote{https://www.bibliotheek.nl/} If this is not the case:
    \begin{enumerate}
        \item Check whether the language of the book is listed on the website of a Dutch reading community website.\footnote{https://www.hebban.nl/}
        \item If step (a) is also not conclusive: check whether more information on the author is available, for instance on a personal website where we can find the original titles.
        \item In case there is no reliable information available on the original language of a book, we discard the book (category 3: no label)
    \end{enumerate}
    \item Book titles with Dutch as their original language are annotated with the label `1' (category 1). Books that were written in a language other than Dutch were annotated with the label `0' (category 2).%
\end{enumerate}

\paragraph{Special cases}
An interesting annotation case regards books from bilingual authors who learned Dutch at a later age, such as Kader Abdolah. In our current guidelines, we do not take this into account specifically; if originally written in Dutch, these books are annotated with category 1.
We note that books that were translated to Dutch were not all originally written in English: other source languages in the data set include German, French and Spanish. %

\vspace{0.5em}

\appendix{{\bf D \hspace{1em} Book-level MTLD comparison of APE and tailored reranking (top-k sampling)}}

\begin{figure*}[h]
    \centering
    \includegraphics[width=0.9\textwidth]{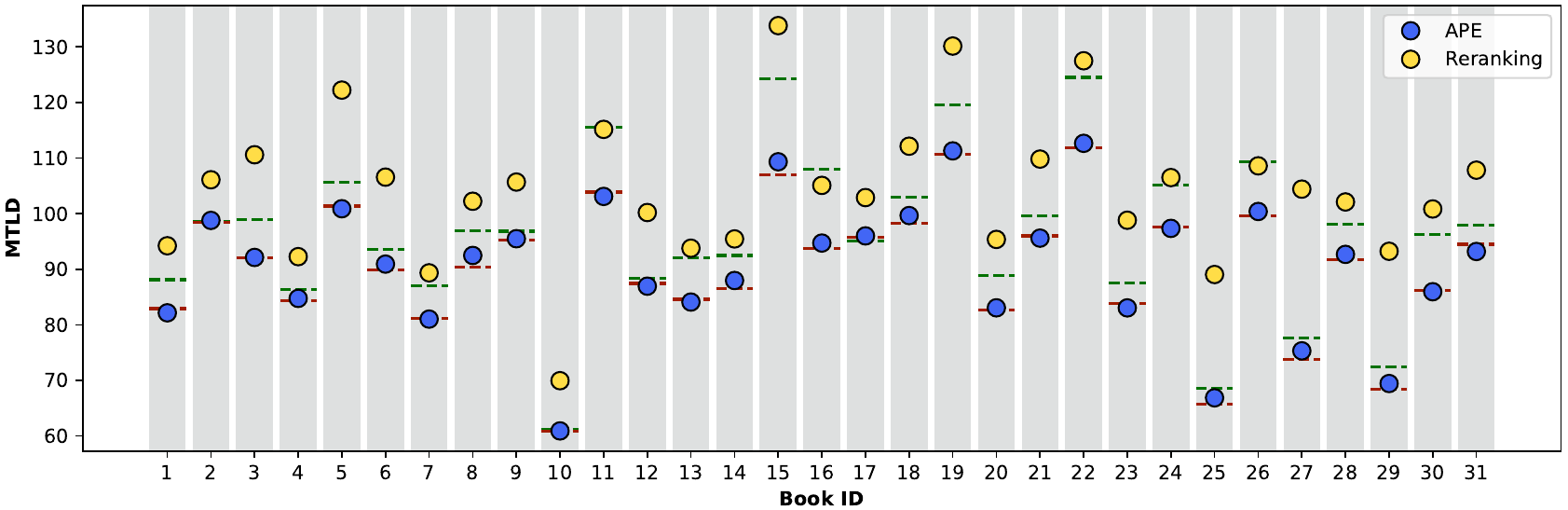}
    \caption{MTLD scores for APE and tailored reranking with top-k sampling, with on the y-axis the MTLD score for each book in our test set (x-axis).}
\end{figure*}

\appendix{{\bf E \hspace{1em} Lexical diversity according to ranks}}
\label{app:lexdiv_ranks}

\begin{figure*}[h]
    \includegraphics[width=0.33\textwidth]{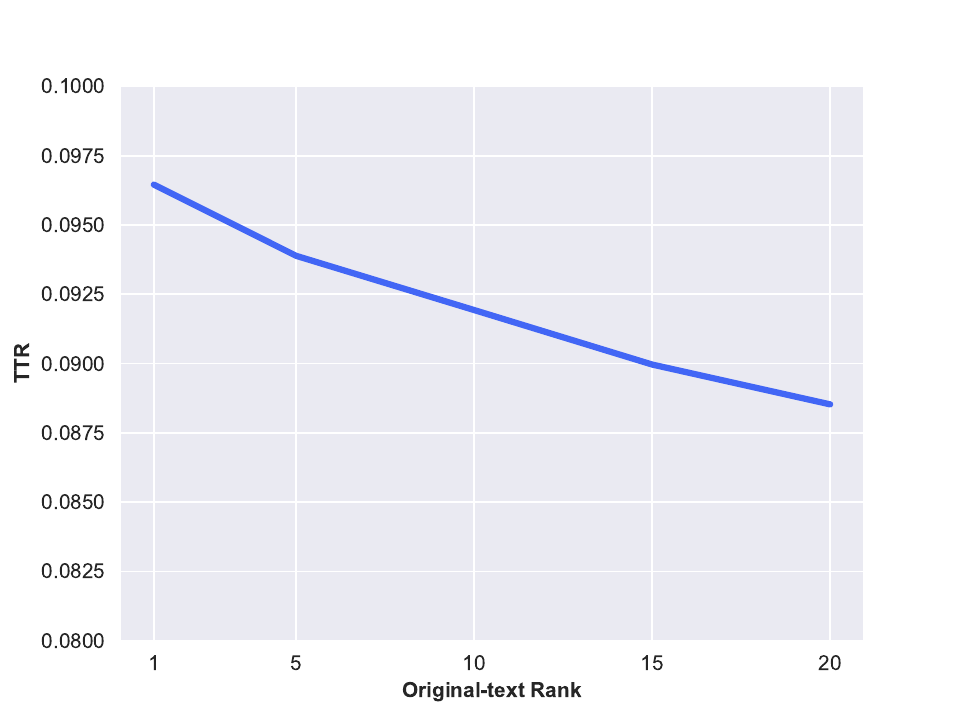}
    \includegraphics[width=0.33\textwidth]{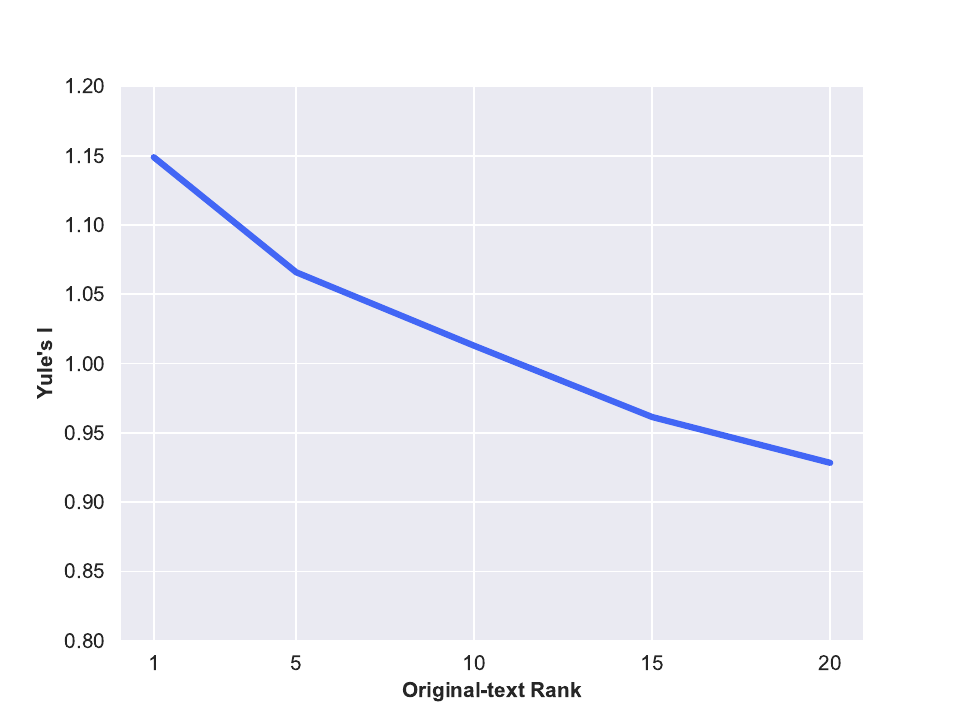}
    \includegraphics[width=0.33\textwidth]{images/flex_demo/mtld_bs20.pdf}
    \caption{TTR, MTLD and Yule's I according to original-text rank, where a higher rank represents smaller original-text probability.}
\end{figure*}

\end{document}